\documentclass[10pt]{article} 
\usepackage[nonatbib,preprint]{tmlr_modified}
\usepackage[numbers,sort&compress]{natbib}

\usepackage{multirow}
\usepackage{subcaption}
\usepackage{xspace}
\usepackage{colortbl}
\usepackage{graphicx}
\usepackage[inline]{enumitem}
\usepackage{adjustbox}

\newcommand{\textcode}[1]{{\fontfamily{cmtt}\selectfont #1}\xspace}

\newcommand{\TechGPTFourProbDesc}{\ensuremath{\textsc{GPT4Hints-Base}}}
\newcommand{\TechGPTFourOutput}{\ensuremath{\textsc{GPT4Hints-IO}}}
\newcommand{\TechGPTFourRepair}{\ensuremath{\textsc{GPT4Hints-IOFix}}}
\newcommand{\TechOurs}{\ensuremath{\textsc{GPT4Hints-GPT3.5Val}}}
\newcommand{\TechTutor}{\ensuremath{\textsc{TutorHints}}}

\newcommand{\DIntroPython}{\ensuremath{\textsc{BasicAlgo}}}
\newcommand{\DIntroDSRegEx}{\ensuremath{\textsc{DataRegex}}}
\newcommand{\DIntroDSAnalysis}{\ensuremath{\textsc{DataAnalysis}}}

\newcommand{\DIntroPythontextbf}{B{\scriptsize \textbf{ASIC}}A{\scriptsize \textbf{LGO}}}
\newcommand{\DIntroDSRegExtextbf}{D{\scriptsize \textbf{ATA}}R{\scriptsize \textbf{EGEX}}}
\newcommand{\DIntroDSAnalysistextbf}{D{\scriptsize \textbf{ATA}}A{\scriptsize \textbf{NALYSIS}}}

\newcommand{\PGCD}{\textsc{GCD}}
\newcommand{\PFibonacci}{\textsc{Fibonacci}}
\newcommand{\PDivisors}{\textsc{DivisorsDiv3}}
\newcommand{\PPalindrome}{\textsc{Palindrome}}
\newcommand{\PMergeStrings}{\textsc{MergeStrs}}

\newcommand{\task}{\ensuremath{\mathcal{T}}}
\newcommand{\buggyprog}{\ensuremath{\mathcal{P}_{\textnormal{b}}}}
\newcommand{\fixedprog}{\ensuremath{\mathcal{P}_{\textnormal{f}}}}
\newcommand{\failingIO}{$\omega$}
\newcommand{\explanation}{\ensuremath{\mathcal{X}}}
\newcommand{\hint}{\ensuremath{\mathcal{H}}}
\newcommand{\trials}{\textit{k}}
\newcommand{\ChatGPT}{{{ChatGPT}}}
\newcommand{\GPTFour}{\ensuremath{\textsc{GPT-4}}}
\newcommand{\GPTThreePFive}{\ensuremath{\textsc{GPT-3.5}}}

\definecolor{ExpHighlight}{rgb}{0.8, 0.1, 0.1}
\definecolor{mygreen}{rgb}{0,0.6,0}
\definecolor{mygray}{rgb}{0.5,0.5,0.5}
\definecolor{mymauve}{rgb}{0.58,0,0.82}
\definecolor{CodeHighlight}{rgb}{1,1,0.6}
\definecolor{CodeGray}{rgb}{0.8,0.8,0.8}
\definecolor{CodeDarkGray}{rgb}{0.6,0.6,0.6}
\definecolor{InputOutput}{rgb}{0.764,0.345,0.009}
\definecolor{promptinputcolor}{rgb}{0.58,0,0.82}
\definecolor{promptheadercolor}{rgb}{0.471,0.318,0.663}

\newcommand{\ourscellcolor}{green!15}

\newcommand{\promptheader}[1]{{\large{\textcolor{promptheadercolor}{\textbf{#1}}}}}
\newcommand{\promptinput}[1]{{\textcolor{promptinputcolor}{\textcode{#1}}}}

\usepackage{arydshln}  

\usepackage{listings}
\makeatletter
\let\old@lstKV@SwitchCases\lstKV@SwitchCases
\def\lstKV@SwitchCases#1#2#3{}
\makeatother
\usepackage{lstlinebgrd}
\makeatletter
\let\lstKV@SwitchCases\old@lstKV@SwitchCases

\lst@Key{numbers}{none}{%
    \def\lst@PlaceNumber{\lst@linebgrd}%
    \lstKV@SwitchCases{#1}%
    {none:\\%
     left:\def\lst@PlaceNumber{\llap{\normalfont
                \lst@numberstyle{\thelstnumber}\kern\lst@numbersep}\lst@linebgrd}\\%
     right:\def\lst@PlaceNumber{\rlap{\normalfont
                \kern\linewidth \kern\lst@numbersep
                \lst@numberstyle{\thelstnumber}}\lst@linebgrd}%
    }{\PackageError{Listings}{Numbers #1 unknown}\@ehc}}
\makeatother
\usepackage{hyperref}
\usepackage{url}

\title{Automating Human Tutor-Style Programming Feedback: Leveraging GPT-4 Tutor Model for Hint Generation and GPT-3.5 Student Model for Hint Validation}

\author{\name Tung Phung \email mphung@mpi-sws.org\\
        \addr Max Planck Institute for Software Systems
        \vspace{-2mm}
        \AND
        \name Victor-Alexandru P{\u a}durean \email vpadurea@mpi-sws.org\\
        \addr Max Planck Institute for Software Systems
        \vspace{-2mm}
        \AND
        \name Anjali Singh \email singhanj@umich.edu\\
        \addr University of Michigan 
        \vspace{-2mm}
        \AND      
        \name Christopher Brooks\thanks{These authors are listed in alphabetical order. Correspondence to: Adish Singla <\texttt{adishs@mpi-sws.org}>.}\email brooksch@umich.edu\\
        \addr University of Michigan 
        \vspace{-2mm}
        \AND      
        \name Jos{\'e} Cambronero\footnotemark[1] \email jcambronero@microsoft.com\\
        \addr Microsoft      
        \vspace{-2mm}
        \AND      
        \name Sumit Gulwani\footnotemark[1] \email sumitg@microsoft.com\\
        \addr Microsoft       
        \vspace{-2mm}
        \AND
        \name Adish Singla\footnotemark[1] \email adishs@mpi-sws.org\\
        \addr Max Planck Institute for Software Systems
        \vspace{-2mm}
        \AND      
        \name Gustavo Soares\footnotemark[1] \email gsoares@microsoft.com\\
        \addr Microsoft
        \vspace{-2mm}
      }

\def\month{MM}  
\def\year{YYYY} 
\def\openreview{\url{https://openreview.net/forum?id=XXXX}} 

\begin{document}

\maketitle

\begin{abstract}
Generative AI and large language models hold great promise in enhancing programming education by automatically generating individualized feedback for students. We investigate the role of generative AI models in providing human tutor-style programming hints to help students resolve errors in their buggy programs. Recent works have benchmarked state-of-the-art models for various feedback generation scenarios; however, their overall quality is still inferior to human tutors and not yet ready for real-world deployment. In this paper, we seek to push the limits of generative AI models toward providing high-quality programming hints and develop a novel technique, \TechOurs{}. As a first step, our technique leverages \GPTFour{} as a ``tutor'' model to generate hints -- it boosts the generative quality by using symbolic information of failing test cases and fixes in prompts. As a next step, our technique leverages \GPTThreePFive{}, a weaker model, as a ``student'' model to further validate the hint quality -- it performs an automatic quality validation by simulating the potential utility of providing this feedback. We show the efficacy of our technique via extensive evaluation using three real-world datasets of Python programs covering a variety of concepts ranging from basic algorithms to regular expressions and data analysis using \emph{pandas} library.
\end{abstract}

\vspace{-2mm}
\section{Introduction}
\vspace{-1mm}
\label{sec.introduction}
Generative AI and large language models (LLMs) have the potential to drastically improve the landscape of computing and programming education by powering next-generation educational technologies. This potential lies in the advanced capabilities of state-of-the-art models---like OpenAI's \GPTFour{}~\cite{GPT4} and \ChatGPT{} (based on \GPTThreePFive{})~\cite{ChatGPT}---to automatically generate high-quality personalized content and feedback for students~\cite{DBLP:journals/corr/abs-2303-12712,icer23poster_ChatGPT_pythonprog,icer23poster_ChatGPT_visualprog}. A series of recent works have already shown us sparks of their capabilities for various programming education scenarios, including generating new programming assignments~\cite{DBLP:conf/icer/SarsaDH022,neurtasksyn}, providing code explanations~\cite{macneil23sigcse,DBLP:conf/icer/SarsaDH022}, repairing buggy programs~\cite{zhang2022repairing,edm23-pyfixv}, enhancing programming-error-messages~\cite{leinonen23sigcse,edm23-pyfixv}, and acting as pair programmer~\cite{CopilotWeb,DBLP:journals/corr/abs-2210-14306}. 

\looseness-1In this paper, we investigate the role of LLMs in providing human tutor-style programming hints to help students resolve errors in their buggy programs. More concretely, given a programming task and a student's buggy program, we want to generate natural language hints to help the student resolve bug(s) and make progress, inspired by how a human tutor would give pedagogical feedback. With the current scale of enrollments in introductory programming courses~\cite{mirhosseini23sigcse}, it has become infeasible for human tutors to promptly provide individualized feedback to students, thereby motivating the need to develop automatic feedback generation techniques. To this end, we aim to leverage generative AI and LLMs for automating human tutor-style programming feedback to support students' learning and reduce human tutors' workload.

\looseness-1Recent works have studied state-of-the-art LLMs for generating various forms of programming feedback for students, including detailed explanations about bugs or single-sentence hints~\cite{icer23poster_ChatGPT_pythonprog,leinonen23sigcse,edm23-pyfixv}. Despite promising initial results, the overall quality of feedback generated by LLMs is substantially inferior to that of human tutors and not yet ready for deployment in real-life classroom settings. For instance, a recent benchmark study in \cite{icer23poster_ChatGPT_pythonprog} evaluated \GPTFour{} in generating hints for buggy programs on introductory Python programming tasks and assessed its quality performance using expert annotations -- \GPTFour{}'s performance in terms of hints quality is only about $60\%$ in contrast to human tutors's performance of over $90\%$. This performance gap between \GPTFour{} vs. human tutors can be attributed to several factors, as discussed next. First, state-of-the-art models still struggle with symbolic reasoning and program execution abilities crucial for understanding the underlying bugs and possible student misconceptions~\cite{DBLP:journals/corr/abs-2303-12712,icer23poster_ChatGPT_pythonprog,icer23poster_ChatGPT_visualprog,DBLP:journals/corr/abs-2302-04023}. Second, these models also suffer from hallucination issues and the generated feedback text---even though seemingly plausible---may contain inaccurate information that could have detrimental effects on students' learning~\cite{DBLP:journals/corr/abs-2302-04023,DBLP:journals/corr/abs-2309-00029,DBLP:conf/icer/LiHFZZK22}. Third, these models still lack a calibration mechanism to decide whether the generated content is of high quality or not~\cite{edm23-pyfixv}; in particular, they are unable to do a human tutor-style reasoning from a student's perspective and judge if the generated feedback would likely help the student.
%
%
\begin{figure}[t!]
    \centering

    \begin{minipage}{\linewidth}
    {
        \begin{subfigure}{\linewidth}
        {
            \begin{tabular}{|p|p|}
                \hline
                \multicolumn{1}{|p{0.485\linewidth}::}{
                    {\small \input{figs/illustration_palindrome_p6/content_obfuscated/task_part1}}
                } & 
                \multicolumn{1}{p{0.485\linewidth}|}{
                    {\small \input{figs/illustration_palindrome_p6/content_obfuscated/task_part2}}
                }\\
                \hline
            \end{tabular}
            \caption{Description of the programming task}
            \label{fig.illustration_palindrome_p6.task}
        }
        \end{subfigure}
    }
    \end{minipage}
    \begin{minipage}{\linewidth}
    {
        \begin{subfigure}{0.35\linewidth}
        {
            \centering
            \begin{tabular}{|p{0.93\linewidth}|}
                \hline
                \multicolumn{1}{|p{1.0\linewidth}|}
                {
                    \centering
                     \scalebox{0.985}{
                         \renewcommand{\arraystretch}{1.6}
\begin{lstlisting}[basicstyle=\fontsize{7.0}{6.82}\ttfamily]
#User function Template for python3
class Solution:
  def reverse(self, b, e, S):
    while b < e:
      S[b], S[e] = S[e], S[b]
      e = e - 1
      b = b + 1
    return S

  def isPalindrome(self, S):
    S = list(S)
    beg = 0
    end = len(S) - 1
    rev = self.reverse(beg,end,S)
    if S == rev:
      return 1
    return 0

#{ Driver Code Starts
#Initial Template for Python 3
if __name__ == '__main__':
  T=int(input())
  for i in range(T):
    S = input()
    ob = Solution()
    answer = ob.isPalindrome(S)
    print(answer)
# } Driver Code Ends
\end{lstlisting}
                     }
                }\\
                \hline
            \end{tabular}
            \caption{Student's buggy program}				
            \label{fig.illustration_palindrome_p6.buggy}
        }
        \end{subfigure}
        \hfill
        \begin{subfigure}{0.397\linewidth}
        {
            \centering
             \begin{tabular}{|p{0.93\linewidth}|}
                \hline
                \multicolumn{1}{|p{1.0\linewidth}|}
                {
                    \centering
                     \scalebox{0.985}{
                         \renewcommand{\arraystretch}{1.6}
\begin{lstlisting}[basicstyle=\fontsize{7.0}{6.82}\ttfamily, linebackgroundcolor={ \ifnum\value{lstnumber}=14\color{CodeHighlight}\fi}]
#User function Template for python3
class Solution:
  def reverse(self, b, e, S):
    while b < e:
      S[b], S[e] = S[e], S[b]
      e = e - 1
      b = b + 1
    return S

  def isPalindrome(self, S):
    S = list(S)
    beg = 0
    end = len(S) - 1
    rev = self.reverse(beg,end,S.copy())
    if S == rev:
      return 1
    return 0

#{ Driver Code Starts
#Initial Template for Python 3
if __name__ == '__main__':
  T=int(input())
  for i in range(T):
    S = input()
    ob = Solution()
    answer = ob.isPalindrome(S)
    print(answer)
# } Driver Code Ends
\end{lstlisting}
                     }
                }\\
                \hline
            \end{tabular}
            \caption{Fixed program}				
            \label{fig.illustration_palindrome_p6.fixed}
        }
        \end{subfigure}
        \hfill
        \begin{subfigure}{0.18\linewidth}
        {
            \centering
             \begin{tabular}{|p{1\linewidth}|}
                \hline
                \multicolumn{1}{|p{1\linewidth}|}
                {                                   
                    {\small \input{figs/illustration_palindrome_p6/content_obfuscated/testcase}}
                }\\
                \\
                \\
                \\
                \\
                \\
                \\
                \\
                \\
                \\
                \\
                \\
                \\
                \hline
            \end{tabular}
            \caption{Failing test case}				
            \label{fig.illustration_palindrome_p6.testcase}
        }
        \end{subfigure}
    }
    \\
    \end{minipage}
    \begin{minipage}{\linewidth}
    {
        \begin{subfigure}{0.56\linewidth}
        {
            \begin{tabular}{|p{1\linewidth}|}
                \hline
                \multicolumn{1}{|p{1\linewidth}|}
                {                                 
                    {\small \input{figs/illustration_palindrome_p6/content_obfuscated/GPTEnhanced_explanation}}
                }\\
                \hline
            \end{tabular}
            \caption{Detailed explanation}				
            \label{fig.illustration_palindrome_p6.GPTEnhanced_explanation}
        }
        \end{subfigure}
        \hfill
        \begin{subfigure}{0.22\linewidth}
        {
            \begin{tabular}{|p{1\linewidth}|}
                \hline
                \multicolumn{1}{|p{1\linewidth}|}
                {                                   
                    {\small \input{figs/illustration_palindrome_p6/content_obfuscated/GPTEnhanced_hint}}
                }\\
                \hline
            \end{tabular}
            \caption{Single-sentence hint}				
            \label{fig.illustration_palindrome_p6.GPTEnhanced_hint}
        }
        \end{subfigure}
        \hfill
        \begin{subfigure}{0.15\linewidth}
        {
             \begin{tabular}{|p{1\linewidth}|}
                \hline
                \multicolumn{1}{|p{1\linewidth}|}{} \\
                \multicolumn{1}{|p{1\linewidth}|}
                {
                    \vspace{0.5mm}
                    \centering
                    \includegraphics[height=1.15cm]{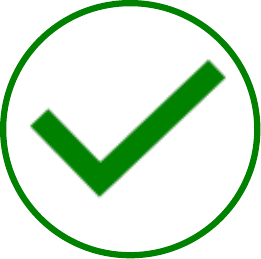}
                    \vspace{0.8mm}                    
                }\\
                \multicolumn{1}{|p{1\linewidth}|}{} \\
                \hline
            \end{tabular}
            \caption{Validation}				
            \label{fig.illustration_palindrome_p6.validation}
        }
        \end{subfigure}
    }
    \end{minipage}
    
    \caption{
         \looseness-1Illustrative example showcasing \TechOurs{} for the \PPalindrome{} problem shown in \textbf{(a)} from the \DIntroPython{} dataset. \textbf{(b)} shows a real-world buggy program. \textbf{(c)} shows a fixed program generated by the technique in an intermediate step, and \textbf{(d)} shows a test case where the buggy program fails to produce the correct output. \textbf{(e)} shows a detailed explanation generated by the technique that is used later in the validation stage. \textbf{(f)} shows the generated feedback (a single-sentence hint). \textbf{(g)} highlights that the validation stage of the technique \emph{successfully accepted} the generated feedback as high-quality and suitable for sharing with the student.
    }
    \vspace{-2mm}
    \label{fig.illustration_palindrome_p6}
\end{figure}

\subsection{Our Approach and Contributions}
In this paper, we seek to push the limits of generative AI and state-of-the-art LLMs toward providing high-quality programming hints. Given a base model, this would require improving the model's abilities at input-level by developing better prompting strategies~\cite{DBLP:conf/nips/Wei0SBIXCLZ22}, at output-level by developing mechanisms to validate the generated content~\cite{edm23-pyfixv,DBLP:journals/corr/abs-2304-05128,DBLP:journals/corr/abs-2303-17651}, or at model-level itself by fine-tuning (when considering open-source models~\cite{DBLP:journals/corr/abs-2307-09288}). In our work, we consider OpenAI's \GPTFour{}~\cite{GPT4} as the base model---the latest model presumably with over a trillion parameters---as it has shown to drastically improve existing models across various programming education scenarios~\cite{icer23poster_ChatGPT_pythonprog}.

\begin{figure}[t!]
    \centering

    \begin{minipage}{\linewidth}
    {
        \begin{subfigure}{\linewidth}
        {
            \begin{tabular}{|p|p|}
                 \hline
                \multicolumn{1}{|p{0.681\linewidth}::}{
                    {\small \input{figs/illustration_pandasanalysis_p28/content_obfuscated/task_part1}}
                } & 
                \multicolumn{1}{p{0.285\linewidth}|}{
                    {\small \input{figs/illustration_pandasanalysis_p28/content_obfuscated/task_part2}}
                }\\
                \hline
            \end{tabular}
            \caption{Description of the programming task}				
            \label{fig.illustration_pandasanalysis_p28.task}
        }
        \end{subfigure}
    }
    \end{minipage}
    \begin{minipage}{\linewidth}
    {
        \begin{subfigure}{0.36\linewidth}
        {
             \begin{tabular}{|p{1\linewidth}|}
                \hline
                \multicolumn{1}{|p{1\linewidth}|}
                {
                    \centering
                    \scalebox{1.0}{
                        \renewcommand{\arraystretch}{1.6}
\begin{lstlisting}[basicstyle=\fontsize{6.6}{6.85}\ttfamily, escapechar=@]
def chickenpox_by_sex():
  import pandas as pd
  import numpy as np
  df = pd.read_csv('assets/data.csv', index_col=0)
  male = df[df['SEX'] == 1]
  vac_m = male[male['P_NUMVRC'] == 1]
  cpox_m = vac_m[vac_m['HAD_CPOX'] == 1]
  no_m = vac_m[vac_m['HAD_CPOX'] == 2]
  total_cpox_m = cpox_m['SEX'].count()
  total_nocpox_m = no_m['SEX'].count()
  m = total_cpox_m/total_nocpox_m

  @\textcolor{CodeDarkGray}{ ... code omitted to save space ... }@

  ans={"male":  m,
       "female": f}

  return ans
\end{lstlisting}
                    }
                }\\
                \hline
            \end{tabular}
            \caption{Student's buggy program}				
            \label{fig.illustration_pandasanalysis_p28.buggy}
        }
        \end{subfigure}
        \hfill
        \begin{subfigure}{0.36\linewidth}
        {
             \begin{tabular}{|p{1\linewidth}|}
                \hline
                \multicolumn{1}{|p{1\linewidth}|}
                {
                    \centering
                    \scalebox{1.0}{
                         \renewcommand{\arraystretch}{1.6}
\begin{lstlisting}[basicstyle=\fontsize{6.6}{6.85}\ttfamily, linebackgroundcolor={ \ifnum\value{lstnumber}=6\color{CodeHighlight}\fi}, escapechar=@]
def chickenpox_by_sex():
  import pandas as pd
  import numpy as np
  df = pd.read_csv('assets/data.csv', index_col=0)
  male = df[df['SEX'] == 1]
  vac_m = male[male['P_NUMVRC'] >= 1]
  cpox_m = vac_m[vac_m['HAD_CPOX'] == 1]
  no_m = vac_m[vac_m['HAD_CPOX'] == 2]
  total_cpox_m = cpox_m['SEX'].count()
  total_nocpox_m = no_m['SEX'].count()
  m = total_cpox_m/total_nocpox_m

  @\textcolor{CodeDarkGray}{ ... code omitted to save space ... }@

  ans={"male":  m,
       "female": f}

  return ans
\end{lstlisting}
                    }
                }\\
                \hline
            \end{tabular}
            \caption{Fixed program}				
            \label{fig.illustration_pandasanalysis_p28.fixed}
        }
        \end{subfigure}
        \hfill
        \begin{subfigure}{0.21\linewidth}
        {
             \begin{tabular}{|p{1\linewidth}|}
                \hline
                \multicolumn{1}{|p{1\linewidth}|}
                {                                   
                    {\small \input{figs/illustration_pandasanalysis_p28/content_obfuscated/testcase_reduced}}
                }\\
                \\
                \hline
            \end{tabular}
            \caption{Failing test case}				
            \label{fig.illustration_pandasanalysis_p28.testcase}
        }
        \end{subfigure}
    }
    \\
    \end{minipage}
    \begin{minipage}{\linewidth}
    {
        \begin{subfigure}{0.59\linewidth}
        {
             \begin{tabular}{|p{1\linewidth}|}
                \hline
                \multicolumn{1}{|p{1\linewidth}|}
                {                                   
                    {\small \input{figs/illustration_pandasanalysis_p28/content_obfuscated/GPTEnhanced_explanation}}
                }\\
                \hline
            \end{tabular}
            \caption{Detailed explanation}				
            \label{fig.illustration_pandasanalysis_p28.GPTEnhanced_explanation}
        }
        \end{subfigure}
        \hfill
        \begin{subfigure}{0.217\linewidth}
        {
             \begin{tabular}{|p{1\linewidth}|}
                \hline
                \multicolumn{1}{|p{1\linewidth}|}
                {                                   
                    {\small \input{figs/illustration_pandasanalysis_p28/content_obfuscated/GPTEnhanced_hint}}
                }\\
                \\
                \hline
            \end{tabular}
            \caption{Single-sentence hint}				
            \label{fig.illustration_pandasanalysis_p28.GPTEnhanced_hint}
        }
        \end{subfigure}
        \hfill
        \begin{subfigure}{0.12\linewidth}
        {
             \begin{tabular}{|p{1\linewidth}|}
                \hline
                \multicolumn{1}{|p{1\linewidth}|}{} \\
                \multicolumn{1}{|p{1\linewidth}|}
                {
                    \vspace{0.85mm}
                    \centering
                    \includegraphics[height=1.15cm]{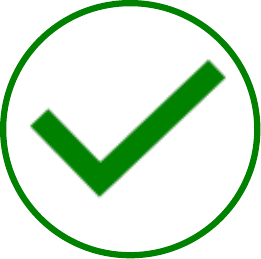}
                    \vspace{0.5mm}                    
                }\\
                \multicolumn{1}{|p{1\linewidth}|}{} \\
                \hline
            \end{tabular}
            \caption{Validation}				
            \label{fig.illustration_pandasanalysis_p28.validation}
        }
        \end{subfigure}
    }
    \end{minipage}
    \vspace{-1mm}
    \caption{
        Similar to Figure~\ref{fig.illustration_palindrome_p6}, this example showcases \TechOurs{} on a buggy program from the \DIntroDSAnalysis{} dataset.
    }
    \label{fig.illustration_pandasanalysis_p28}
    \vspace{-1mm}
\end{figure}

\looseness-1We develop a novel technique, \TechOurs{}, to provide human tutor-style high-quality programming hints. Our technique leverages the \GPTFour{} model in the role of a ``tutor'' to generate hints and boosts the generative quality at the input level by prompting it with symbolic information of failing test cases and fixed programs. At the output level, it further validates the hint quality by leveraging the \GPTThreePFive{} model as a ``student'' to simulate the potential utility of providing this feedback to human students. This validation step is designed to provide a quality assurance layer and decides whether the generated feedback should be provided to the human student or not -- thereby trading off \emph{coverage} (how many students are given automatic feedback) and \emph{precision} (quality of the given feedback). We show the efficacy of our technique by conducting an extensive evaluation using three real-world datasets of Python programs covering a variety of concepts ranging from writing basic algorithms to regular expressions and data analysis using \emph{pandas}~\cite{mckinney2011pandas}. Figures~\ref{fig.illustration_palindrome_p6}~and~\ref{fig.illustration_pandasanalysis_p28} showcase \TechOurs{} on two different buggy programs.\footnote{When presenting these illustrative examples in this paper, we slightly obfuscate the students' buggy programs to avoid showing exact real-world programs. We do so by altering variable names and formatting conventions while preserving the original bugs exactly the same, as considered in related works~\cite{edm23-pyfixv,icer23poster_ChatGPT_pythonprog}. Accordingly, if needed, we apply the same adjustments to the generated output to maintain consistency with these alterations.}
More broadly, our work makes the following contributions in leveraging generative AI and LLMs for computing and programming education:
\begin{enumerate}[label={\Roman*.},leftmargin=10pt, parsep=-1pt]
    \item We showcase the utility of prompting the models with symbolic information, such as failing test cases and fixed programs, to enhance their reasoning abilities about the underlying bugs crucial for providing high-quality hints.
    \item We showcase the utility of using LLMs in a flipped role as a ``student'' model to simulate the potential effect of feedback on real human students. Our results highlight that using a weaker model (\GPTThreePFive{}, instead of \GPTFour{}) provides better validation of programming hints from \GPTFour{}. This flipped role opens up new opportunities in utilizing generative AI for in-context student modeling for automatic assessments, learning analytics, and simulations.
    \item Our technique achieves a precision of around $95\%$ (reaching the quality of human tutors in our evaluation) while maintaining a high coverage of over $70\%$ across three real-world Python programming datasets.\footnote{\url{https://github.com/machine-teaching-group/lak2024_GPT4Hints-GPT3.5Val}  \label{footnote.githublink}}
\end{enumerate}

\vspace{-4mm}
\subsection{Related Work}
\vspace{-0.5mm}
\looseness-1\emph{Feedback generation for programming education.} Prior to recent developments in generative AI and LLMs, the research on feedback generation for programming education had primarily focused on fixing buggy programs because of challenges in automatically generating natural language explanations~\cite{singh2013automated,DBLP:conf/pldi/GulwaniRZ18}. A parallel line of research explored crowdsourcing approaches to obtain explanations provided by other students/tutors~\cite{DBLP:conf/lats/HeadGSSFDH17}. Our work builds on recent developments in leveraging LLMs for generating programming feedback~\cite{icer23poster_ChatGPT_pythonprog,leinonen23sigcse,edm23-pyfixv,DBLP:journals/corr/abs-2307-00150}, in particular, motivated by recent survey~\cite{icer23poster_ChatGPT_pythonprog} that highlighted a substantial gap in \GPTFour{}'s performance in terms of hints quality w.r.t. human tutors. Another closely related work is \cite{edm23-pyfixv} that proposed \textsc{PyFiXV} technique for generating high-precision feedback for syntax errors. \textsc{PyFiXV} has a run-time feedback validation mechanism by leveraging OpenAI's Codex-Edit model~\cite{codexedit} at varying temperatures as a ``student'' model. Inspired by \cite{edm23-pyfixv}, we also leverage an LLM-based ``student'' model to perform validation. However, the validation mechanism used in \textsc{PyFiXV} is not directly applicable to our setting as it is designed only for syntax errors that substantially simplify the validation process; crucially, \TechOurs{} is designed to provide feedback for any types of errors a student might encounter, including errors related to the program's time complexity.

\looseness-1\emph{Enhancing a model's generative performance.} A series of recent works have focused on enhancing the generative performance of a base model in a \emph{black-box setting}, given the high monetary or computational costs involved in fine-tuning state-of-the-art models (in fact, the latest OpenAI's \GPTFour{} model doesn't have public APIs for fine-tuning). These works operate either at the input level by developing better prompting strategies~\cite{DBLP:conf/nips/Wei0SBIXCLZ22} or at the output level by analyzing and correcting the generated content~\cite{edm23-pyfixv,DBLP:journals/corr/abs-2304-05128,DBLP:journals/corr/abs-2303-17651}. At the output level enhancements, \emph{Self-Debugging}~\cite{DBLP:journals/corr/abs-2304-05128} and \emph{Self-Refine}~\cite{DBLP:journals/corr/abs-2303-17651} are two recently proposed methods that enable an LLM to analyze and correct its output automatically. Another recent work in \cite{DBLP:journals/corr/abs-2306-09896} introduced the concept of \emph{Self-Repair} that showed substantial performance gains when allowing an LLM to repair its output by receiving feedback from a more powerful LLM or expert. The key intuition behind the validation mechanism in \TechOurs{} differs from these works and is more related to~\cite{edm23-pyfixv} discussed above---we utilize another LLM as a  ``student'' model to simulate the potential effect of feedback on real human students.

\emph{Integration of generative AI in educational sites.} 
There has also been increasing interest in integrating generative AI and LLMs in educational sites. For instance, Khanmigo~\cite{Khanmigo} by Khan Academy and Q-Chat by Quizlet~\cite{Q-Chat} are AI-powered systems based on OpenAI's GPT models. These recent developments also serve as our motivation to develop principled techniques that can generate high-quality feedback. Overall, we see our work as complementary to these systems and believe that the proposed techniques can be useful in further improving the performance of these systems.
\vspace{-1mm}
\section{Problem Setup}
\vspace{-1mm}
\label{sec.problem}
\looseness-1\textit{Programming task and student's buggy program as input.} We start with a programming task \task{} and a buggy program \buggyprog{}. A task \task{}, such as shown in Figures ~\ref{fig.illustration_palindrome_p6.task} and ~\ref{fig.illustration_pandasanalysis_p28.task}, is represented by a textual description of the programming problem. Additionally, this description encompasses all requisite information essential for problem solving, such as expected algorithm complexity and any constraints on input, as applicable. In cases where the task necessitates interaction with an external file, \task{}
should also contain all pertinent information of that file crucial for solving the problem, such as the file's format or structure. 
\buggyprog{}, as illustrated in Figures ~\ref{fig.illustration_palindrome_p6.buggy} and ~\ref{fig.illustration_pandasanalysis_p28.buggy}, is an unsuccessful attempt of the student to solve \task{}. This program fails to pass at least one of the test cases in the test suite for \task{}. In general, \buggyprog{} may contain one or multiple errors, spanning various error types including syntax and semantic errors.

\textit{Tutor-style hint as output and quality assessment.} Given \task{} and \buggyprog{}, we aim to generate a human tutor-style natural language hint \hint{} as feedback to aid the student in understanding and resolving the programming error. We assess the quality of generated feedback along four quality attributes following the rubric used in \cite{icer23poster_ChatGPT_pythonprog}. All attributes are binary, with a value of $1$ being better. \emph{HCorrect} captures whether the generated hint provides correct information for resolving issues in the student's buggy program. \emph{HInformative} captures whether the generated hint provides useful information to help the student resolve bug(s); this attribute is set to $0$ by default when the hint is incorrect. \emph{HConceal} captures that the information in the generated hint is not too detailed, so the student would also have to reason about implementing the fixes; this attribute is set to $0$ by default when the hint is incorrect. \emph{HComprehensible} captures whether the generated hint is easy to understand, presented in a readable format, and doesn't contain redundant information. In our evaluation, human experts (evaluators) assess the quality of generated hints along these four attributes. We measure the overall quality of the generated hint by \emph{HOverall} that takes the value of $1$ (good quality) if all the four quality attributes are satisfied and otherwise $0$ (bad quality).

\looseness-1\textit{Performance metrics and objective.} Next, we describe the overall performance metrics used to evaluate a feedback generation technique. For a given student's buggy program \buggyprog{}, we seek to design techniques that generate feedback and also decide whether the generated feedback is suitable for sharing with the student. Similar to \cite{edm23-pyfixv}, we measure the performance of a technique using two metrics: (i) \emph{Coverage} measuring the percentage number of times the generated feedback is provided to the student; (ii) \emph{Precision} measuring the percentage number of times the provided feedback is of good quality w.r.t. the \emph{HOverall} quality introduced above. In our experiments, we will compute these metrics on a dataset comprising a set of students' buggy programs. Our goal is to design feedback generation techniques with high precision, which is imperative before deploying such techniques in classrooms. In particular, we aim to develop techniques that achieve a precision level of human tutors while maintaining an effective trade-off between precision and coverage.
\vspace{-1.5mm}
\section{Our Technique: \TechOurs}
\begin{figure}[t!]
    \centering
    \includegraphics[width=\linewidth]{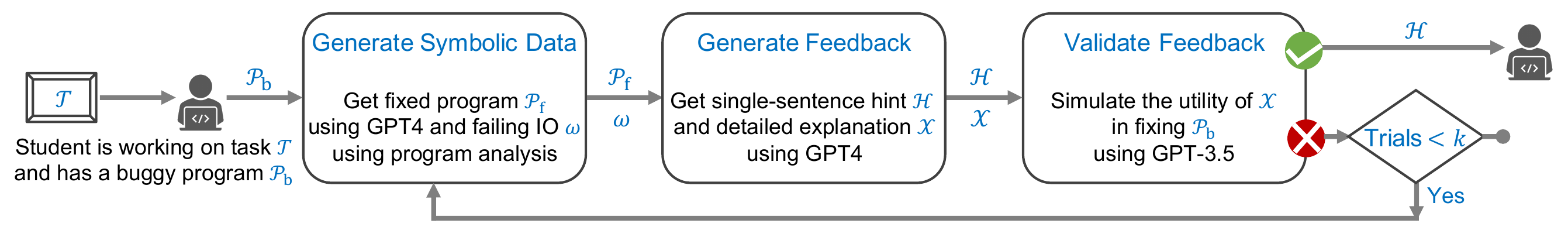}
    \caption{
        Illustration of different stages in \TechOurs{}'s feedback generation process.
    }
    \label{fig.technique_overview}
\end{figure}

\vspace{-1.5mm}
This section gives details about our proposed technique, namely \TechOurs{}, which leverages and improves upon generative AI models for feedback generation. Figure \ref{fig.technique_overview} shows an overview of our technique. In essence, \TechOurs{} employs \GPTFour{} as a simulated ``tutor'' model for generating feedback and \GPTThreePFive{} as a simulated ``student'' model for feedback validation. In Section \ref{sec.method.symbolic}, we describe two types of symbolic information that are helpful for generating feedback and how to obtain them; in Section \ref{sec.method.generation}, we describe the process of feedback generation augmented with this symbolic information. Subsequently, in Section \ref{sec.method.validation}, we introduce a novel validation mechanism aiming to elevate the precision of the delivered feedback while maintaining a high level of coverage.

\subsection{Stage-1: Generate Symbolic Data}
\label{sec.method.symbolic}
\looseness-1\textit{Overview and intuition.} As discussed in Section \ref{sec.introduction}, there remains a notable performance gap between state-of-the-art generative AI models and human tutors regarding hint generation. One key factor contributing to this disparity is the inability to do symbolic reasoning and program execution. \GPTFour{} lacks the capability to execute the given code to retrieve an output, which could help it gain deeper understanding of the underlying bugs. To mitigate this gap, we employ external tools to execute programs and extract useful symbolic information. We then supply this information to \GPTFour{} for feedback generation. Our approach centers on leveraging two categories of symbolic data: failing test cases and fixed programs.

\textit{Input/output for a failing test case.} To highlight the error in the buggy program \buggyprog{}, we provide \GPTFour{} with a test case for which \buggyprog{} fails to produce the expected output. To acquire this test case, we run \buggyprog{} on the existing test suite given for the corresponding task \task{}. The first test case in which \buggyprog{} fails is selected. We denote the triplet comprising this input, the output generated by \buggyprog{}, and the expected output, as \failingIO{} and include it in the prompt for feedback generation.

\looseness-1\textit{Fixed program.} The fixed program, denoted as \fixedprog{}, is generated using \GPTFour{}, employing a procedure adapted from the work in \cite{edm23-pyfixv}. To be more specific, we initiate the process by requesting the model to produce $10$ independent fixed programs. For this purpose, we include \task{} and \buggyprog{} in the prompt\footnote{The prompt used here has the same format as shown in Figure~\ref{fig.technique_prompts} (third prompt).} to ask for 10 outputs (each output contains a fixed program) with the hyperparameter \textit{temperature} set to $0.5$. 
Then, from this set of $10$, we take the programs that pass the test suite for \task{} and among them, identify \fixedprog{} as the one with the smallest token-edit distance w.r.t. \buggyprog{}.
To compute the token-edit distance between two programs, we first tokenize them using Pygments library \cite{pygments} and then calculate the Levenshtein edit distance based on the tokenized strings. If \fixedprog{} is found, we include it in the prompt for feedback generation. If, however, none of the generated programs is correct, we opt to exclude this symbolic information from the prompt.

\subsection{Stage-2: Generate Feedback}
\label{sec.method.generation}
\textit{Overview and intuition.} In this stage, we aim to obtain a human tutor-style hint \hint{} as feedback to be given to the student, as previously mentioned in Section~\ref{sec.problem}. In addition to our request for a hint \hint{} from \GPTFour{}, we also ask for a detailed explanation, denoted as \explanation{}, for the bugs in \buggyprog{}. The reason to ask for this explanation draws inspiration from Chain-of-Thought \cite{DBLP:conf/nips/Wei0SBIXCLZ22}, an established method renowned for enhancing the reasoning capabilities of LLMs. The essence of the Chain-of-Thought approach lies in encouraging LLMs to explain their thought process meticulously, step by step, prior to presenting the final output. Within the specific context of hint generation, we allow the model to elaborate its reasoning through \explanation{} before coming up with the concise single-sentence hint \hint{}, which is essentially an abstracted version of the explanation. Furthermore, \explanation{} will also play a pivotal role in the subsequent feedback validation stage, which will be elaborated upon in Section \ref{sec.method.validation}.

\looseness-1\textit{Prompt for feedback generation.} In Figure~\ref{fig.technique_prompts} (first prompt), we provide our prompt for generating feedback. This prompt comprises the problem description for \task{}, the buggy program \buggyprog{}, the symbolic information as extracted from the previous stage, and a request for an explanation \explanation{} along with a hint \hint{}. To get a response from \GPTFour{}, we use this prompt while configuring the hyperparameter \textit{temperature} to $0$, indicating our preference for the most probable answer. All other hyperparameters are kept at their default settings. Following this, \explanation{} and \hint{} are then extracted automatically from the output.

\subsection{Stage-3: Validate Feedback}
\label{sec.method.validation}
\textit{Overview and intuition.} This validation stage aims to enhance the precision of the feedback provided to the student. It is worth noting that despite the inclusion of augmented symbolic information in the prompt, the hint generated in Stage-2 may not always align with the desired quality criteria outlined in Section \ref{sec.problem}. To mitigate this issue, we introduce a validation mechanism that adds a run-time quality assurance layer and decides whether the generated feedback is suitable for sharing with the student. The key idea behind this validation mechanism is to leverage an additional AI model as a ``student'' model to simulate the potential utility of providing this feedback to human students. More concretely, we seek to evaluate the quality of feedback by assessing its impact on the simulated students' ability to fix the bugs. If the simulated students find it easier to fix \buggyprog{} with the help of the feedback, then the feedback is deemed high-quality and can be subsequently provided to the real student. In terms of the ``student'' model, we use a weaker model \GPTThreePFive{}, instead of \GPTFour{}. The key intuition is that a weaker model provides a better differential effect in quantifying the utility of feedback in fixing the buggy program; moreover, we use the ``student'' model at a high temperature to add further stochasticity in the process of fixing the program.\footnote{We refer the reader to recent results in \cite{icer23poster_ChatGPT_pythonprog,DBLP:conf/iticse/SavelkaABSS23} to see the performance of different GPT-based models across various programming education scenarios.}. Furthermore, we will use the detailed explanation \explanation{} (instead of the single-sentence hint \hint{}) to assess the utility of feedback for fixing the bugs. In our evaluation (Section~\ref{sec.experiments.results} and Figure~\ref{fig.experiments_results_variations}), we will demonstrate the effectiveness of these design choices.

\textit{Two prompts for validation.} Figure~\ref{fig.technique_prompts} (second and third prompts) illustrates the two prompts used by the feedback validation mechanism. Both prompts essentially instruct the ``student'' model (\GPTThreePFive{}) to fix \buggyprog{}. 
The primary distinction lies in the fact that, in contrast to the third (standard) prompt, the second (augmented) prompt additionally incorporates the explanation \explanation{}. More concretely, the third (standard) prompt is the same as the prompt used in Stage-1 when generating a fixed program; the second (augmented) prompt puts emphasis on the detailed explanation to serve as an instruction for the ``student'' model when fixing the program.
For each prompt, we ask \GPTThreePFive{} to generate a set of $n=10$ independent outputs (the \textit{temperature} is set to $0.5$, similar to in Stage-1), effectively utilizing \GPTThreePFive{} in the role of $10$ simulated students. We shall denote the number of correct output programs resulting from the standard prompt as $n_1$, and the number of correct output programs resulting from the augmented prompt as $n_2$. The correctness of a program is determined by its ability to pass the whole test suite for the corresponding task \task{}. Next, we explain how we use these quantities for feedback validation.

%
\begin{figure}[tp!]
    \centering
    \begin{subfigure}[b]{1\textwidth}
        \centering
        \scalebox{0.85}{
        \setlength\tabcolsep{6pt}
        \renewcommand{\arraystretch}{1.3}
\begin{tabular}{||p{1.1\linewidth}||}
    \hline
    \multicolumn{1}{||c||}{\promptheader{Prompt to Generate Feedback}} \\ 
    I'm working on a Python programming problem. The current program below is not working well. Can you help by giving a hint?
    \newline
    \newline
    Problem description:
    \newline
    \promptinput{\{problem\_description\}}
    \newline
    \newline
    Failing test case:
    \newline
    \promptinput{\{failing\_test\_case\}}
    \newline
    \newline
    Buggy program:
    \newline
    \promptinput{\{buggy\_program\}}
    \newline
    \newline
    The fixed program of the buggy program above:
    \newline
    \promptinput{\{fixed\_program\}}
    \newline
    \newline    
    (1) Can you describe the bug(s) in this program and the required fixes?
    \newline
    \looseness-1(2) Can you provide a concise single-sentence hint about one bug in this program? The hint should not be too detailed as I want to think about the fixes by myself. However, the hint should not be too abstract, as I need some help.
    \\
    \hline
\end{tabular}
        }
    \end{subfigure}
    \\
    \vspace{6mm}
    %
    \begin{subfigure}[b]{1\textwidth}
        \centering
        \scalebox{0.85}{
        \setlength\tabcolsep{6pt}
        \renewcommand{\arraystretch}{1.3}
\begin{tabular}{||p{1.1\linewidth}||}
    \hline
    \multicolumn{1}{||c||}{\promptheader{Prompt to Validate Feedback: (i) Fixing the Program with Explanation}} \\ 
    I'm working on a Python programming problem. The current program below is not working well. Can you help in fixing this program according to a given explanation of the bug(s)? Below I first provide the problem description, the current buggy program, and then the explanation of the bug(s).
    \newline
    \newline
    Problem description:
    \newline
    \promptinput{\{problem\_description\}}
    \newline
    \newline
    Buggy program:
    \newline
    \promptinput{\{buggy\_program\}}
    \newline
    \newline
    The explanation of the bug(s) in the buggy program:
    \newline
    \promptinput{\{explanation\}}
    \newline
    \newline    
    If anything in the explanation above is incorrect or too confusing, please say ``Explanation is bad.'' and stop. If all the reasoning in the explanation above is correct and easy to understand, then please fix the buggy program according to the explanation above. In this case, note that the explanation above may not cover all bugs (if there are multiple bugs) in the buggy program, so you need to think to resolve the remaining bugs by yourself.
    \\
    \hline
\end{tabular}

        }
    \end{subfigure}
    \\
    \vspace{6mm}
    %
    \begin{subfigure}[b]{1\textwidth}
        \centering
        \scalebox{0.85}{
        \setlength\tabcolsep{6pt}
        \renewcommand{\arraystretch}{1.3}
\begin{tabular}{||p{1.1\linewidth}||}
    \hline
    \multicolumn{1}{||c||}{\promptheader{Prompt to Validate Feedback: (ii) Fixing the Program without Explanation}} \\ 
    I'm working on a Python programming problem. The current program below is not working well. Can you help in fixing this program with as few changes as possible? Below I first provide the problem description and then the current buggy program.
    \newline
    \newline
    Problem description:
    \newline
    \promptinput{\{problem\_description\}}
    \newline
    \newline
    Buggy program:
    \newline
    \promptinput{\{buggy\_program\}}
    \newline
    \newline
    Can you fix the above buggy program? Make sure that you make minimal possible changes needed to fix the program.    
    \\
    \hline
\end{tabular}
        }
    \end{subfigure}
    \\
    %
    
    \caption{
        Prompts employed by \TechOurs{} for feedback generation (first) and feedback validation (second and third).
    }
    \label{fig.technique_prompts}
\end{figure}

\textit{Validation threshold rules.} Our main idea for validation is that good feedback should help students find it easier to fix the buggy program than without it. Thus, the primary rule for feedback validation is to have $\frac{n_2}{n} \ge \frac{n_1}{n}$. Nonetheless, in situations where $n_1$ assumes particularly low values, e.g., $n_1=0$ or $n_1=1$, this condition becomes less stringent, and any feedback, regardless of its quality, may pass the validation. To address this, we incorporate an additional requirement to ensure that $\frac{n_2}{n}$ attains a sufficient level independently. This is achieved through the inclusion of the following condition: $\big(\frac{n_2}{n} \ge \alpha \big) \lor \big(\frac{n_2}{n} \ge \frac{n_1}{n} + \beta \big)$, where we instantiate $\alpha$ as $0.50$ and $\beta$ as $0.25$. In other words, we require the ratio of correct output programs generated with the help of the explanation to either exceed a certain fixed threshold (i.e., $\frac{n_2}{n} \ge 0.5$) or be substantially higher than the ratio of correct output programs generated without the explanation (i.e., $\frac{n_2}{n} \ge \frac{n_1}{n} + 0.25$), or both. Consequently, our final validation mechanism approves a feedback instance only when the following condition holds true: 
$ \Big(\big(\frac{n_2}{n} \ge \frac{n_1}{n}\big) \land \big(\big(\frac{n_2}{n} \ge 0.50\big) \lor \big(\frac{n_2}{n} \ge \frac{n_1}{n} + 0.25\big)\big)\Big) $,
and rejects it otherwise. In our experiments (Section~\ref{sec.experiments}), we will also compare the performance of different variants of threshold rules.

\textit{Multiple trials.} When the validation mechanism rejects a feedback instance, it is not provided to the human student. While this is expected to boost the precision metric, it could also lead to a significant drop in the coverage metric~\cite{edm23-pyfixv}. Given the stochasticity of the generation and validation processes, we introduce an additional layer to the overall process to boost the coverage while ensuring high precision. More concretely, if a feedback instance is rejected, we restart the process, including acquiring symbolic information, generating hints, and the subsequent validation. We maintain this iterative cycle until either a generated feedback instance is approved by the validation mechanism or a predefined maximum number of iterations, denoted as \trials{}, is attained (we set \trials{} $= 3$). After \trials{} trials, if none of the feedback instances pass validation, we terminate this outer loop and will not provide any feedback to the human student. When deploying our technique in real-world classroom settings, where no automatic feedback is being provided, a human tutor could step in and take over the work of providing feedback to the student.

\vspace{-2mm}
\section{Experimental Evaluation}
\vspace{-3mm}
\label{sec.experiments}
In this section, we evaluate our technique, \TechOurs{}, across three datasets spanning different domains of introductory Python programming. We assess \TechOurs{} in comparison to baselines such as \GPTFour{} and human tutors. Furthermore, we compare our validation with various alternative variants. In our experiments, we use OpenAI's \GPTFour{} (model=\textit{gpt-4-0613}) as the ``tutor'' model and ChatGPT based on \GPTThreePFive{} (model=\textit{gpt-3.5-turbo-0613}) as the ``student'' model unless otherwise stated.

\begin{figure}[t!]
    \centering
    \scalebox{0.75}{
    \setlength\tabcolsep{1.2pt}
    \renewcommand{\arraystretch}{1.6}
        \begin{tabular}{l||l|l|l}
            \hline
            \multicolumn{1}{c||}{\textbf{Properties}} &
            \multicolumn{1}{c|}{{\DIntroPythontextbf{}}} &
            \multicolumn{1}{c|}{{\DIntroDSRegExtextbf{}}} &
            \multicolumn{1}{c}{{\DIntroDSAnalysistextbf{}}} \\
            \hline
            \multicolumn{1}{l||}{Number of programming tasks} &
            \multicolumn{1}{c|}{5} &
            \multicolumn{1}{c|}{1} &
            \multicolumn{1}{c}{1} \\
            \multicolumn{1}{l||}{Number of buggy programs} &
            \multicolumn{1}{c|}{25} &
            \multicolumn{1}{c|}{24} &
            \multicolumn{1}{c}{30} \\
            \multicolumn{1}{l||}{Average lines of student code} &
            \multicolumn{1}{c|}{10.7} &
            \multicolumn{1}{c|}{2.2} &
            \multicolumn{1}{c}{12.1} \\
            \multicolumn{1}{l||}{Task's objective} &
            \multicolumn{1}{l|}{Write an algorithm in Python} &
            \multicolumn{1}{l|}{Fix a regular expression in Python} &
            \multicolumn{1}{l}{Perform data analysis in Python}  \\
            \multicolumn{1}{l||}{Domain and concepts} &
            \multicolumn{1}{l|}{Python syntax, basic algorithms} &
            \multicolumn{1}{l|}{Regular expressions, information extraction} &
            \multicolumn{1}{l}{\emph{pandas} library, data analysis} \\
            \hline
        \end{tabular}
    }
    \caption{
        Overview of the datasets used in this work. See Section \ref{sec.experiments.datasets} for details.
    }
    \vspace{-2mm}
    \label{fig.experiments_datasets}
\end{figure}



\vspace{-1mm}
\subsection{Datasets}
\label{sec.experiments.datasets}
\vspace{-1mm}
\looseness-1To comprehensively assess the techniques' performance across diverse domains within introductory programming education, we use three datasets representing different types of learning objectives, as summarized in Figure \ref{fig.experiments_datasets}. All datasets consist of students' Python buggy programs. Below, we provide a detailed description of each of these datasets.

\looseness-1The first dataset, \DIntroPython{}, was introduced in \cite{icer23poster_ChatGPT_pythonprog}. It covers five popular introductory Python problems, and for each problem, there are five corresponding buggy programs. The problems capture a diverse set of basic programming concepts and include the following: \PGCD{} (finding the greatest common divisor of two given numbers), \PFibonacci{} (generating the list of Fibonacci numbers up to a given value), \PDivisors{} (counting the number of divisors that divide $3$ of a given number), \PPalindrome{} (checking whether a given string is palindrome or not), and \PMergeStrings{} (merging two given strings alternatively). The buggy programs come from different users on the \textit{geeksforgeeks.org} platform \cite{geeksforgeeks}, and capture a variety of bug types and code lengths. Figures~\ref{fig.illustration_palindrome_p6}~and~\ref{fig.illustration_merge_p6} show two examples of buggy programs with bugs related to misconception regarding the mutability of lists and a mistake regarding the ordering of the merging strings.

The second dataset, \DIntroDSRegEx{}, comes from an introductory data science programming course. This course is a part of an online Master's degree program in applied data science; students enrolling in the course are required to have basic Python programming and statistics knowledge. We examine the second exercise from the first assignment of the course, which requires students to use regular expressions to extract information from a text file. In particular, the text file contains people's names and their corresponding grades; the students need to fix a given buggy function so that it correctly reads the file, matches a regular expression, captures and returns a list of people who got a grade of B.\footnote{For \GPTFour{}, instead of giving it the file, we describe the file format in the prompt; the description is provided as part of our implementation (see  Footnote~\ref{footnote.githublink}).} To solve the problem, students need knowledge of basic regular expression concepts such as wildcard characters, grouping, look around, and quantification. This dataset contains $24$ buggy submissions, each from a unique student. For each student, if there are multiple buggy submissions, we take only the median submission w.r.t to submission times to include in the dataset. Some common types of bugs are mishandling of grouping (Figure~\ref{fig.illustration_regex_p144}), returning names of all people, and returning only people's last names. It is worth noting that there is only one test case in the test suite for this problem; this is in contrast to algorithmic problems, such as the ones in \DIntroPython{}, in which the test suites usually comprise a large number of input/output cases.

\looseness-1The third dataset, \DIntroDSAnalysis{}, is from the second exercise of the second assignment in the same data science course. By that time, the students learnt to use data manipulation libraries such as \textit{pandas} to load, filter, and extract meaningful information from data-frames. For this problem, the students are given a \textit{csv} format file that contains a data-frame, a $252$-page data guide PDF,\footnote{The data guide is meant to exercise students on extracting relevant information. Typically, students would search the PDF using keywords such as 'chickenpox' to spot relevant columns needed. For GPT-4, we extract and provide in the prompt a short summary describing the relevant columns; the summary is provided as part of our implementation (see Footnote \ref{footnote.githublink}).} a problem description, and a function signature. The students need to complete the given empty function to compute the ratios of vaccinated children who contracted chickenpox versus those who were vaccinated but did not contract chickenpox, separated by sex. To solve this problem, besides the basic Python syntax, the students also need to know how to select and use relevant libraries (such as \textit{pandas}), understand and search for relevant information from the extensive data guide, and deal with missing data. To form this third dataset, we sample $30$ buggy programs using the same procedure as used for second dataset. Some bugs in the dataset are: mis-filtering of data (Figure \ref{fig.illustration_pandasanalysis_p28}), misreading of the requirements and computing a wrong ratio, and forgetting to handle or wrongly handling of missing values.

\subsection{Baselines and Variants of Our Technique}
\label{sec.experiments.techniques}

\textit{Baseline GPT-4 and human tutors.}
As our first baseline, we employ \GPTFour{} in a straightforward manner by presenting it with the task description and the buggy program in the prompt to generate feedback. The format of the prompt closely resembles that depicted in Figure 4 (first prompt), albeit without the inclusion of additional symbolic information. The second baseline employs human tutors with experience in Python programming and tutoring, which serves as the gold standard for our technique to match. In our experiments, two human tutors are employed to give hints independently. From here on, we refer to these baselines as \TechGPTFourProbDesc{} and \TechTutor{}, respectively.

\textit{Variants of our technique without validation.}
As mentioned previously, we introduce two additional types of symbolic information into our prompt for feedback generation. These additions consist of a failing test case and a fixed program, given that a correct fixed program can be produced (see Section~\ref{sec.method.symbolic}). 
Accordingly, we have formulated two variant techniques:
\begin{enumerate*}[label=(\roman*)]
\item \TechGPTFourOutput{} involves enhancing \TechGPTFourProbDesc{} by incorporating the failing test case into the prompt;
\item \TechGPTFourRepair{} integrates both of these types of symbolic information into the prompt.
\end{enumerate*}
Note that neither of these techniques employ any validation, i.e., the generated feedback is always deemed suitable for sharing.

\looseness-1\textit{Variations of validation stage in our technique.}
Next, we will consider variants of \TechOurs{} in terms of the validation stage. First, we look at the role of multiple trials when a feedback instance fails validation. We compare our technique with a variant where there is only a single trial (i.e., \trials{} $= 1$). Second, we examine the performance when \GPTFour{} is used as the simulated ``student'' model instead of \GPTThreePFive{}. Third, we investigate the case wherein the generated single-sentence hint, instead of the detailed explanation, is utilized in the validation process. Fourth and last, we vary the threshold rule used for validation. In this regard, there are three variations: 
$\big(\frac{n_2}{n} \ge \alpha \big)$,
where $n_1$ is not considered in the rule;
$\big(\big(\frac{n_2}{n} \ge \frac{n_1}{n} \big) \land \big(\frac{n_2}{n} \ge \alpha \big)\big)$ where $\beta$ is not considered in the rule; 
$\big(\frac{n_2}{n} \ge \frac{n_1}{n} \big)$ where $\alpha$ and $\beta$ are not considered in the rule.

\subsection{Evaluation Procedure}
\label{sec.experiments.procedure}
As discussed in Section~\ref{sec.problem}, we employ human experts (evaluators) to assess the quality of generated feedback. More concretely, two human evaluators independently rated the feedback generated by techniques along the quality attributes as introduced in Section~\ref{sec.problem}.\footnote{Similar to \cite{icer23poster_ChatGPT_pythonprog}, these two human evaluators are same as two human tutors employed in the \TechTutor{} technique. When evaluating \TechTutor{} technique, an evaluator does not assess their own feedback produced while acting as a tutor.} Then, given the ratings from each evaluator, we compute precision and coverage (based on the overall feedback quality \textit{HOverall}).\footnote{In addition, we also asked the evaluators to rate on \textit{ECorrect}, a binary attribute capturing the correctness of the detailed explanation \explanation{}. Further analysis regarding this additional attribute will be discussed in Section~\ref{sec.experiments.results} and Figure~\ref{fig.experiments_results_allmetrics_pythyon5x5}.} Finally, for each technique and dataset, we aggregate across evaluators and report averaged results as \emph{mean} (\emph{stderr}). We obtained Cohen's kappa reliability value $0.65$ indicating \textit{substantial agreement} between evaluators \cite{cohen1960coefficient}. Next, we elaborate on our experimental results.

\begin{figure}[t!]
    \centering
    \scalebox{0.78}{
    \setlength\tabcolsep{7.5pt}
    \renewcommand{\arraystretch}{1.2}
    \begin{tabular}{l||cc|cc|cc}
        \hline
        \multicolumn{1}{c||}{\textbf{Technique}} & \multicolumn{2}{c|}{\textbf{\DIntroPython}} & \multicolumn{2}{c|}{\textbf{\DIntroDSRegEx}} & \multicolumn{2}{c}{\textbf{\DIntroDSAnalysis}}\\
         & \multicolumn{1}{c|}{Precision} & \multicolumn{1}{c|}{Coverage} & \multicolumn{1}{c|}{Precision} & \multicolumn{1}{c|}{Coverage} & \multicolumn{1}{c|}{Precision} & \multicolumn{1}{c}{Coverage} \\  
        \hline
        \TechGPTFourProbDesc & $66.0~(2.0)$ & $100.0$ & $85.4~(2.1)$ & $100.0$ & $78.3~(5.0)$ & $100.0$ \\
        \TechGPTFourOutput & $72.0~(4.0)$ & $100.0$ & $85.4~(2.1)$ & $100.0$ & $85.0~(5.0)$ & $100.0$ \\
        \TechGPTFourRepair & $82.0~(2.0)$ & $100.0$ & $91.7~(4.2)$ & $100.0$ & $93.3~(3.3)$ & $100.0$ \\
        \TechTutor & $92.0~(4.0)$ & $100.0$ & $91.7~(4.2)$ & $100.0$ & $91.7~(8.3)$ & $100.0$ \\
        \hline
        {\cellcolor{\ourscellcolor}\TechOurs{}}  \qquad \qquad & {\cellcolor{\ourscellcolor}$94.7~(0.0)$} & {\cellcolor{\ourscellcolor}$\ \ \ \ \ \ \ \ \ 76.0~(0.0)$} & {\cellcolor{\ourscellcolor}$97.6~(2.4)$} & {\cellcolor{\ourscellcolor}$\ \ \ \ \ \ \ \ \ 87.5~(0.0)$} & {\cellcolor{\ourscellcolor}$95.5~(4.5)$} & {\cellcolor{\ourscellcolor}$\ \ \ \ \ \ \ \ \ 73.3~(0.0)$} \\
        \hline
    \end{tabular}
    }
    \caption{
        \looseness-1Results for different techniques on three real-world Python programming datasets. For each technique and dataset, results are averaged across two evaluators and reported as mean (stderr) as per the evaluation procedure in Section~\ref{sec.experiments.procedure}.
        Our technique, \TechOurs{}, performs validation of the generated feedback to achieve a higher quality of the feedback in terms of precision level, thereby trading off precision and coverage.
        Our technique can achieve a precision of around $95\%$ reaching the quality of human tutors while maintaining a high coverage of over $70\%$ across three real-world datasets;  see Section~\ref{sec.experiments.results} for a detailed discussion of results.
    }
    \label{fig.experiments_results_main}
\end{figure}


\begin{figure}[t!]
    \scalebox{0.78}{
        \setlength\tabcolsep{3pt}
        \renewcommand{\arraystretch}{1.45}
        \begin{tabular}{l||cc|cc|cc}
        \hline
        \multicolumn{1}{c||}{\textbf{Variants of Validation Stage}} & \multicolumn{2}{c|}{\textbf{\DIntroPython}} & \multicolumn{2}{c|}{\textbf{\DIntroDSRegEx}} & \multicolumn{2}{c}{\textbf{\DIntroDSAnalysis}}\\
        \multicolumn{1}{c||}{\textbf{in \TechOurs{} }} & \multicolumn{1}{c|}{Precision} & \multicolumn{1}{c|}{Coverage} & \multicolumn{1}{c|}{Precision} & \multicolumn{1}{c|}{Coverage} & \multicolumn{1}{c|}{Precision} & \multicolumn{1}{c}{Coverage} \\  
        \hline
        {\cellcolor{\ourscellcolor}Default with trials $\trials{}=3$, \GPTThreePFive{} student model, use \explanation{}, and} & {\cellcolor{\ourscellcolor}$94.7~(0.0)$} & {\cellcolor{\ourscellcolor}$\ \ 76.0$} & {\cellcolor{\ourscellcolor}$97.6~(2.4)$} & {\cellcolor{\ourscellcolor}$\ \ 87.5$} & {\cellcolor{\ourscellcolor}$95.5~(4.5)$} & {\cellcolor{\ourscellcolor}$\ \ 73.3$} \\
        {\cellcolor{\ourscellcolor}threshold rule $ \Big(\big(\frac{n_2}{n} \ge \frac{n_1}{n}\big)
 \land \big(\big(\frac{n_2}{n} \ge 0.50\big) \lor \big(\frac{n_2}{n} \ge \frac{n_1}{n} + 0.25\big)\big)\Big)$} & {\cellcolor{\ourscellcolor} } & {\cellcolor{\ourscellcolor} } & {\cellcolor{\ourscellcolor} } & {\cellcolor{\ourscellcolor} } & {\cellcolor{\ourscellcolor} } & {\cellcolor{\ourscellcolor} } \\
        \hline 
        Single trial $\trials{}=1$ instead of $\trials{}=3$& $91.7~(0.0)$ & $\ \ 48.0$ & $96.4~(3.6)$ & $\ \ 58.3$ & $94.4~(5.6)$ & $\ \ 60.0$ \\
        \GPTFour{} student model instead of \GPTThreePFive{}{} student model& $84.8~(2.2)$ & $\ \ 92.0$ & $93.5~(2.2)$ & $\ \ 95.8$ & $93.1~(3.4)$ & $\ \ 96.7$ \\
        Using single-sentence hint \hint{} instead of detailed explanation \explanation{} & $89.3~(3.6)$ & $\ \ 56.0$ & $93.5~(2.2)$ & $\ \ 95.8$ & $95.0~(5.0)$ & $\ \ 66.7$ \\
        Threshold rule without considering $n_1$, i.e., $\big(\frac{n_2}{n} \ge 0.50\big)$ & $86.8~(2.6)$ & $\ \ 76.0$ & $91.7~(4.1)$ & $\ \ 100.0$ & $95.5~(4.5)$ & $\ \ 73.3$ \\
        \hline
        Simplified threshold rule without $\beta$, i.e., $ \Big(\big(\frac{n_2}{n} \ge \frac{n_1}{n}\big) \land \big(\frac{n_2}{n} \ge 0.50\big)\Big)$ & $94.1~(0.0)$ & $\ \ 68.0$ & $97.6~(2.4)$ & $\ \ 87.5$ & $95.5~(4.5)$ & $\ \ 73.3$ \\
        Simplified threshold rule without $\alpha,\beta$, i.e., $\big(\frac{n_2}{n} \ge \frac{n_1}{n}\big)$ & $95.2~(0.0)$ & $\ \ 84.0$ & $97.6~(2.4)$ & $\ \ 87.5$ & $92.3~(3.8)$ & $\ \ 86.7$ \\
        \hline
    \end{tabular}
    }
    \caption{
        Comparison of performance between \TechOurs{} and different variants w.r.t the validation stage. The first four variations (single trial, \GPTFour{} student model, using \hint{}, and threshold without considering $n_1$) show how different design choices in our validation stage helps improve precision-coverage trade off. The last two variations with simplified threshold rules shows the robustness of the default threshold rule in terms of $\alpha$ and $\beta$. See Sections~\ref{sec.method.validation}~and~\ref{sec.experiments.results} for further details.
    }
    \label{fig.experiments_results_variations}
\end{figure}

\subsection{Results}
\label{sec.experiments.results}
\textit{Comparison with baselines and human tutors.}
Figure~\ref{fig.experiments_results_main} provides an overview of results,  comparing our technique and baselines. It is evident that \TechGPTFourProbDesc{} exhibits a substantial performance gap when compared to \TechTutor{}. This gap is partially mitigated with the incorporation of failing test cases and fixed programs in the prompt, as seen with \TechGPTFourOutput{} and \TechGPTFourRepair{}, respectively.\footnote{If, for a buggy program, no correct fixed program is obtained (see Section \ref{sec.method.symbolic}), the prompt of \TechGPTFourRepair{} is the same as \TechGPTFourOutput{}'s. The rates at which we obtained at least one correct fix for \DIntroPython{}, \DIntroDSRegEx{}, and \DIntroDSAnalysis{} datasets are $92\%$, $100\%$, and $93\%$, respectively.} Our final technique, \TechOurs{}, consistently achieves precision levels comparable to \TechTutor{}, around $95\%$ across all datasets.\footnote{When comparing \TechOurs{} with other techniques in Figure~\ref{fig.experiments_results_main}, the results are significantly different w.r.t. $\chi^2$ tests~\cite{cochran1952chi2} ($p\leq0.0001$); here, we use contingency tables with two rows (techniques) and four columns (data points are mapped to four possible precision/coverage outcomes).}
Importantly, the trade-off in coverage required to attain such high precision is effective, and our technique maintains a coverage rate exceeding 70\% for all three datasets.
In Figure \ref{fig.experiments_results_allmetrics_pythyon5x5}, we provide 
fine-grained results across different attributes, demonstrating a high correlation between generating a high-quality hint and a correct detailed explanation -- this further justifies why the explanation can be used to validate the hint.

\textit{Comparison with variations of validation stage.}
Figure \ref{fig.experiments_results_variations} shows the performance of different variants in comparison to our technique. Notably, with a single trial (i.e., \trials{} $= 1$), there is a substantial decrease in coverage across all datasets. This result underscores the marked effect of incorporating multiple trials in maintaining a high coverage level. Intriguingly, when we substitute \GPTThreePFive{} with the more advanced model, \GPTFour{}, as the simulated ``student'' model, there is actually a reduction in precision. We observed that \GPTFour{} is worse than \GPTThreePFive{} in terms of achieved precision as it tends to correctly fix the buggy program even if the explanation in the validation prompt is wrong. 
These results highlight that a weaker model (here, \GPTThreePFive{} instead of \GPTFour{}) could be better suited as a simulated ``student'' model.
When using hints instead of explanations for validation, it yields inferior performance in general as the explanation contains more details about the bugs and fixes (thus having a  better differential effect between using the standard and the augmented prompt). Regarding variants of the validation rule, the overall performance remains relatively stable when $\alpha$ and $\beta$ are excluded from the rule, suggesting a robust performance irrespective of specific settings for these hyperparameters. However, a noticeable decline in performance is observed when the relative condition ($\frac{n_2}{n} \ge \frac{n_1}{n}$) is omitted, highlighting its importance in the validation process.

\begin{figure}[t!]
    \centering
    \scalebox{0.78}{
        \setlength\tabcolsep{3.3pt}
        \renewcommand{\arraystretch}{1.2}
        \begin{tabular}{l||ccccc|c|c}
            \hline
            \multicolumn{1}{c||}{\textbf{Method}} & \multicolumn{5}{c|}{\textbf{Hint}} & \multicolumn{1}{c|}{\textbf{Explanation}} & \multicolumn{1}{c}{\textbf{(Hint, Explanation)}}\\
             & \multicolumn{1}{c}{HOverall} & \multicolumn{1}{c}{HCorrect} & \multicolumn{1}{c}{HInformative} & \multicolumn{1}{c}{HConceal} & \multicolumn{1}{c|}{HComprehensible} & \multicolumn{1}{c|}{ECorrect} & \multicolumn{1}{c}{HOverall, ECorrect}\\  
            \hline
            \TechGPTFourProbDesc & $66.0$ & $68.0$ & $66.0$ & $68.0$ & $100.0$ & $58.0$ & 56.0 \\
            \TechGPTFourOutput & $72.0$ & $78.0$ & $74.0$ & $76.0$ & $98.0$ & $66.0$ & 62.0 \\
            \TechGPTFourRepair & $82.0$ & $84.0$ & $82.0$ & $84.0$ & $100.0$ & $82.0$ & 80.0 \\
            \hline
            {\cellcolor{\ourscellcolor}\TechOurs} & {\cellcolor{\ourscellcolor}$94.7$} & {\cellcolor{\ourscellcolor}$94.7$} & {\cellcolor{\ourscellcolor}$94.7$} & {\cellcolor{\ourscellcolor}$94.7$} & {\cellcolor{\ourscellcolor}$100.0$} & {\cellcolor{\ourscellcolor}$91.1$} & {\cellcolor{\ourscellcolor}$92.1$} \\
            \hline
        \end{tabular}
    }
    \caption{
        Fine-grained results w.r.t. evaluation rubric that assesses the quality of generated feedback across different attributes as discussed in Sections~\ref{sec.problem} and \ref{sec.experiments.procedure}. For our technique, these fine-grained results demonstrate a high correlation between generating a high-quality hint and a correct detailed explanation (used in the validation stage).
    }
    \label{fig.experiments_results_allmetrics_pythyon5x5}
\end{figure}

\begin{figure}[t!]
    \centering

    \begin{minipage}{\linewidth}
    {
        \begin{subfigure}{\linewidth}
        {
            \begin{tabular}{|p|p|}
                 \hline
                \multicolumn{1}{|p{0.54\linewidth}::}{
                    {\small \input{figs/illustration_regex_p144/content_obfuscated/task_part1}}
                } & 
                \multicolumn{1}{p{0.43\linewidth}|}{
                    {\small \input{figs/illustration_regex_p144/content_obfuscated/task_part2}}
                }\\
                \hline
            \end{tabular}
            \caption{Description of the programming task}				
            \label{fig.illustration_regex_p144.task}
        }
        \end{subfigure}
    }
    \end{minipage}
    \begin{minipage}{\linewidth}
    {
        \begin{subfigure}{0.382\linewidth}
        {
             \begin{tabular}{|p{1\linewidth}|}
                \hline
                \multicolumn{1}{|p{1\linewidth}|}
                {
                     \scalebox{1.0}{
                         \renewcommand{\arraystretch}{1.6}
\begin{lstlisting}[basicstyle=\fontsize{7.0}{6.4}\ttfamily]
def student_grades():
  import re
  with open ("assets/grades.txt", "r") as f:
    grades=f.read()

  ### FIX CODE BELOW
  re_pattern="(\w+) (\w+)(?=: B)"
  matches=re.findall(re_pattern,grades)
  ### FIX CODE ABOVE

  return matches
\end{lstlisting}
                     }
                }\\
                \hline
            \end{tabular}
            \caption{Student's buggy program}				
            \label{fig.illustration_regex_p144.buggy}
        }
        \end{subfigure}
        \hfill
        \begin{subfigure}{0.382\linewidth}
        {
             \begin{tabular}{|p{1\linewidth}|}
                \hline
                \multicolumn{1}{|p{1\linewidth}|}
                {
                    \centering
                     \scalebox{1.0}{
                         \renewcommand{\arraystretch}{1.6}
\begin{lstlisting}[basicstyle=\fontsize{7.0}{6.4}\ttfamily, linebackgroundcolor={ \ifnum\value{lstnumber}=7\color{CodeHighlight}\fi}]
def student_grades():
  import re
  with open ("assets/grades.txt", "r") as f:
    grades=f.read()

  ### FIX CODE BELOW
  re_pattern="(\w+ \w+)(?=: B)"
  matches=re.findall(re_pattern,grades)
  ### FIX CODE ABOVE

  return matches
\end{lstlisting}
                     }
                }\\
                \hline
            \end{tabular}
            \caption{Fixed program}				
            \label{fig.illustration_regex_p144.fixed}
        }
        \end{subfigure}
        \hfill
        \begin{subfigure}{0.17\linewidth}
        {
             \begin{tabular}{|p{1\linewidth}|}
                \hline
                \multicolumn{1}{|p{1\linewidth}|}
                {                                   
                    {\small \input{figs/illustration_regex_p144/content_obfuscated/testcase_reduced}}
                }\\
                \hline
            \end{tabular}
            \caption{Failing test case}				
            \label{fig.illustration_regex_p144.testcase}
        }
        \end{subfigure}
        \\
    }
    \end{minipage}
    \begin{minipage}{\linewidth}
    {
        \begin{subfigure}{0.58\linewidth}
        {
             \begin{tabular}{|p{1\linewidth}|}
                \hline
                \multicolumn{1}{|p{1\linewidth}|}
                {                                   
                    {\small \input{figs/illustration_regex_p144/content_obfuscated/GPTEnhanced_explanation}}
                }\\
                \hline
            \end{tabular}
            \caption{Detailed explanation}				
            \label{fig.illustration_regex_p144.GPTEnhanced_explanation}
        }
        \end{subfigure}
        \hfill
        \begin{subfigure}{0.22\linewidth}
        {
             \begin{tabular}{|p{1\linewidth}|}
                \hline
                \multicolumn{1}{|p{1\linewidth}|}
                {                                   
                    {\small \input{figs/illustration_regex_p144/content_obfuscated/GPTEnhanced_hint}}
                }\\
                \\
                \hline
            \end{tabular}
            \caption{Single-sentence hint}				
            \label{fig.illustration_regex_p144.GPTEnhanced_hint}
        }
        \end{subfigure}
        \hfill
        \begin{subfigure}{0.125\linewidth}
        {
             \begin{tabular}{|p{1\linewidth}|}
                \hline
                \multicolumn{1}{|p{1\linewidth}|} {} \\
                \multicolumn{1}{|p{1\linewidth}|}
                {
                    \centering
                    \includegraphics[height=1.15cm]{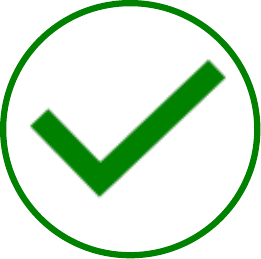}
                }\\
                \multicolumn{1}{|p{1\linewidth}|} {} \\
                \hline
            \end{tabular}
            \caption{Validation}				
            \label{fig.illustration_regex_p144.validation}
        }
        \end{subfigure}
    }
    \end{minipage}
    
    \caption{
        Similar to Figure~\ref{fig.illustration_palindrome_p6}, this example showcases \TechOurs{} on a buggy program from the \DIntroDSRegEx{} dataset.
    }
    \label{fig.illustration_regex_p144}
    \vspace{-4mm}
\end{figure}

\begin{figure}[t!]
    \centering

    \begin{minipage}{\linewidth}
    {
        \begin{subfigure}{\linewidth}
        {
            \begin{tabular}{|p|p|}
                 \hline
                \multicolumn{1}{|p{0.57\linewidth}::}{
                    {\small \input{figs/illustration_merge_p6/content_obfuscated/task_part1}}
                } & 
                \multicolumn{1}{p{0.4\linewidth}|}{
                    {\small \input{figs/illustration_merge_p6/content_obfuscated/task_part2}}
                }\\
                \hline
            \end{tabular}
            \caption{Description of the programming task}				
            \label{fig.illustration_merge_p6.task}
        }
        \end{subfigure}
    }
    \end{minipage}
    \begin{minipage}{\linewidth}
    {
        \begin{subfigure}{0.35\linewidth}
        {
             \begin{tabular}{|p{1\linewidth}|}
                \hline
                \multicolumn{1}{|p{1\linewidth}|}
                {
                     \scalebox{1.0}{
                         \renewcommand{\arraystretch}{1.6}
\begin{lstlisting}[basicstyle=\fontsize{7.0}{6.31}\ttfamily]
#User function Template for python3
class Solution:
  def merge(self, S1, S2):
    l1=len(S1)
    l2=len(S2)
    if l1>l2:
      res=""
      dif=l1-l2
      x=S1[:l2]
      y=S2
      d=[(i,j) for i,j in zip(x,y)]
      for i in d:
        for j in i:
          res+=j
      return res+S1[-dif:]
    elif l1<l2:
      res=""
      dif=l2-l1
      x=S2[:l1]
      y=S1
      d=[(i,j)for i,j in zip(x,y)]
      for i in d:
        for j in i:
          res+=j
      return res+S2[-dif:]
    else:
      res=""
      x=S1[:l2]
      y=S2
      d=[(i,j)for i,j in zip(x,y)]
      for i in d:
        for j in i:
          res+=j
      return res

#{ Driver Code Starts
#Initial Template for Python 3
if __name__ == '__main__':
    t = int(input())
    for _ in range(t):
        S1,S2 = map(str,input().strip().split())
        ob = Solution()
        print(ob.merge(S1, S2))
# } Driver Code Ends
\end{lstlisting}
                     }
                }\\
                \hline
            \end{tabular}
            \caption{Student's buggy program}				
            \label{fig.illustration_merge_p6.buggy}
        }
        \end{subfigure}
        \hfill
        \begin{subfigure}{0.35\linewidth}
        {
             \begin{tabular}{|p{1\linewidth}|}
                \hline
                \multicolumn{1}{|p{1\linewidth}|}
                {
                     \scalebox{1.0}{
                         \renewcommand{\arraystretch}{1.6}
\begin{lstlisting}[basicstyle=\fontsize{7.0}{6.35}\ttfamily, linebackgroundcolor={ \ifnum\value{lstnumber}=15\color{CodeHighlight}\fi \ifnum\value{lstnumber}=19\color{CodeHighlight}\fi \ifnum\value{lstnumber}=20\color{CodeHighlight}\fi \ifnum\value{lstnumber}=25\color{CodeHighlight}\fi \ifnum\value{lstnumber}=28\color{CodeHighlight}\fi}]
#User function Template for python3
class Solution:
  def merge(self, S1, S2):
    l1=len(S1)
    l2=len(S2)
    if l1>l2:
      res=""
      dif=l1-l2
      x=S1[:l2]
      y=S2
      d=[(i,j) for i,j in zip(x,y)]
      for i in d:
        for j in i:
          res+=j
      return res+S1[l2:]
    elif l1<l2:
      res=""
      dif=l2-l1
      x=S1
      y=S2[l1:]
      d=[(i,j)for i,j in zip(x,y)]
      for i in d:
        for j in i:
          res+=j
      return res+S2[l1:]
    else:
      res=""
      x=S1
      y=S2
      d=[(i,j)for i,j in zip(x,y)]
      for i in d:
        for j in i:
          res+=j
      return res

#{ Driver Code Starts
#Initial Template for Python 3
if __name__ == '__main__':
    t = int(input())
    for _ in range(t):
        S1,S2 = map(str,input().strip().split())
        ob = Solution()
        print(ob.merge(S1, S2))
# } Driver Code Ends
\end{lstlisting}
                     }
                }\\
                \hline
            \end{tabular}
            \caption{Fixed program}				
            \label{fig.illustration_merge_p6.fixed}
        }
        \end{subfigure}
        \hfill
        \begin{subfigure}{0.235\linewidth}
        {
             \begin{tabular}{|p{1\linewidth}|}
                \hline
                \multicolumn{1}{|p{1\linewidth}|}
                {                                   
                    {\small \input{figs/illustration_merge_p6/content_obfuscated/testcase}}
                }\\
                \\
                \\
                \\
                \\
                \\
                \\
                \\
                \\
                \\
                \\
                \\
                \\
                \\
                \\
                \\
                \\
                \\
                \\
                \\
                \\
                \\
                \hline
            \end{tabular}
            \caption{Failing test case}				
            \label{fig.illustration_merge_p6.testcase}
        }
        \end{subfigure}
        \\
    }
    \end{minipage}
    \begin{minipage}{\linewidth}
    {
        \begin{subfigure}{0.60\linewidth}
        {
             \begin{tabular}{|p{1\linewidth}|}
                \hline
                \multicolumn{1}{|p{1\linewidth}|}
                {                                   
                    {\small \input{figs/illustration_merge_p6/content_obfuscated/GPTEnhanced_explanation}}
                }\\
                \hline
            \end{tabular}
            \caption{Detailed explanation}				
            \label{fig.illustration_merge_p6.GPTEnhanced_explanation}
        }
        \end{subfigure}
        \hfill
        \begin{subfigure}{0.205\linewidth}
        {
             \begin{tabular}{|p{1\linewidth}|}
                \hline
                \multicolumn{1}{|p{1\linewidth}|}
                {                                   
                    {\small \input{figs/illustration_merge_p6/content_obfuscated/GPTEnhanced_hint}}
                }\\
                \\
                \\
                \hline
            \end{tabular}
            \caption{Single-sentence hint}				
            \label{fig.illustration_merge_p6.GPTEnhanced_hint}
        }
        \end{subfigure}
        \hfill
        \begin{subfigure}{0.125\linewidth}
        {
             \begin{tabular}{|p{1\linewidth}|}
                \hline
                \multicolumn{1}{|p{1\linewidth}|}{} \\
                \multicolumn{1}{|p{1\linewidth}|}{} \\
                \multicolumn{1}{|p{1\linewidth}|}
                {
                    \centering
                    \vspace{-4mm}                    
                    \includegraphics[height=1.15cm]{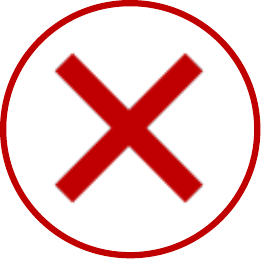}
                    \vspace{1mm}                    
                }\\
                \multicolumn{1}{|p{1\linewidth}|}{} \\
                \hline
            \end{tabular}
            \caption{Validation}				
            \label{fig.illustration_merge_p6.validation}
        }
        \end{subfigure}
    }
    \end{minipage}
    
    \caption{
        Similar to Figure \ref{fig.illustration_palindrome_p6}, this example showcases \TechOurs{} on a buggy program for the \PMergeStrings{} problem from the \DIntroPython{} dataset.
        For this example, the generated detailed explanation and single-sentence hint feedback are not correct (e.g., the explanation suggests fixing the program based on a different slicing strategy, which is not related to the bug in this program). The validation stage of the technique (that evaluates the potential utility of this detailed explanation, cf. Figure~\ref{fig.technique_overview}) \emph{successfully rejected} the generated hint as low-quality and not suitable for sharing with the student. See Section~\ref{sec.experiments.results} for further discussion of results.
    }
    \label{fig.illustration_merge_p6}
    \vspace{-4mm}
\end{figure}

\looseness-1\textit{Qualitative analysis.}
We have included a few illustrative examples to showcase the effectiveness of our technique. Figures \ref{fig.illustration_palindrome_p6}, \ref{fig.illustration_pandasanalysis_p28}, and \ref{fig.illustration_regex_p144} exemplify cases where \TechOurs{} generated high-quality feedback during Stage-2 and then successfully accepted during Stage-3. Conversely, for the scenario in Figure \ref{fig.illustration_merge_p6}, \TechOurs{}'s Stage-2 failed to produce high-quality feedback in all three trials, but Stage-3 successfully rejected all of those low-quality feedback instances. To be more specific, the values of $n_1$ and $n_2$ for the three trials in this case were \{$n_1=8, n_2=0$\}, \{$n_1=6, n_2=0$\}, and \{$n_1=5, n_2=0$\}, respectively. In contrast, in the example shown in Figure \ref{fig.illustration_palindrome_p6}, \TechOurs{}'s Stage-2 generated high-quality feedback during the first trial and Stage-3 subsequently accepted it with values \{$n_1=2, n_2=6$\}. We have provided additional illustrative examples as part of our implementation (see Footnote~\ref{footnote.githublink}).

\vspace{-1mm}
\section{Concluding Discussions}
\vspace{-1mm}
We investigated the role of generative AI and large language models in providing human tutor-style programming hints to help students resolve errors in their buggy programs. In particular, we focused on improving the quality of generated feedback, which is crucial for deployment in real-life classroom settings. We developed a novel technique, \TechOurs{}, that leverages \GPTFour{} as a ``tutor" model to generate hints and \GPTThreePFive{} as a ``student" model to validate the hint quality. This validation step provides a layer of quality assurance by trading off coverage (how many students are given automatic feedback) and precision (quality of the given feedback). We performed an extensive evaluation to showcase the efficacy of our technique on three real-world Python programming datasets, reaching the precision-level of human tutors.

\looseness-1Our work has two important implications for the research community interested in leveraging generative AI and large language models for computing and programming education. First, our results show how we can effectively utilize these models as ``tutor'' by prompting them with symbolic data such as failing test cases. This symbolic data essentially provides in-context information to enhance the reasoning and execution abilities of these models where they typically struggle. Second, our results show how we can utilize these models in a flipped role as ``student'' to simulate the effect of feedback on a real human student. Interestingly, we also showed that a weaker model (\GPTThreePFive{}, instead of \GPTFour{}) serves as a better ``student'' model for validating the effect of feedback generated by \GPTFour{}. This flipped role opens up new opportunities in utilizing generative models as in-context student models for automatic assessments, learning analytics, and simulations.

Next, we discuss some limitations of our current work and ideas to tackle them in the future. 
First, our work involved OpenAI's GPT family of models; it would be useful to evaluate alternate generative models, in particular, open-source variants like Llama-2. Moreover, we utilized the \GPTThreePFive{} model at a higher temperature to simulate the potential utility of providing feedback; it would be interesting to investigate how to employ different LLMs to better simulate diverse student behaviors.
Second, our work didn't leverage historical data on a given problem when generating hints, e.g., hints provided by human tutors for previous students' buggy attempts on a problem. It would be important to develop techniques that can leverage this data, e.g., by fine-tuning these open-source variants to generate better-quality hints. 
Third, our evaluation considered small datasets comprising a total of $79$ buggy programs; it would be useful to scale up the studies by considering larger-scale datasets.
Fourth, we focused only on Python programming education; it would be interesting to conduct a similar study for other programming languages and other domains beyond programming.
Fifth, our evaluation only considered expert-based annotations and didn't involve students; it would be important to conduct studies with students to evaluate techniques from their perspectives.
%

\textbf{Acknowledgments.} Funded/Co-funded by the European Union (ERC, TOPS, 101039090). Views and opinions expressed are however those of the author(s) only and do not necessarily reflect those of the European Union or the European Research Council. Neither the European Union nor the granting authority can be held responsible for them.

\clearpage
\bibliographystyle{unsrt}
\bibliography{main}

\end{document}